\documentclass[10pt,twocolumn,letterpaper]{article}

\usepackage[accsupp]{axessibility}  

\usepackage{wacv}
\usepackage{times}
\usepackage{epsfig}
\usepackage{graphicx}
\usepackage{amsmath}
\usepackage{amssymb}
\usepackage{booktabs}

\usepackage{multirow}
\usepackage{multicol}
\usepackage[linesnumbered,ruled,vlined]{algorithm2e}

\usepackage{subcaption}
\def\wacvPaperID{229} 

\wacvalgorithmstrack   

\wacvfinalcopy 

\ifwacvfinal
\usepackage[breaklinks=true,bookmarks=false]{hyperref}
\else
\usepackage[pagebackref=true,breaklinks=true,colorlinks,bookmarks=false]{hyperref}
\fi
\usepackage{color}

\usepackage[capitalize]{cleveref}
\crefname{section}{Sec.}{Secs.}
\Crefname{section}{Section}{Sections}
\Crefname{table}{Table}{Tables}
\crefname{table}{Tab.}{Tabs.}

\newcommand{\bz}{\mathbf{z}}
\newcommand{\bp}{\mathbf{p}}

\newcommand{\calD}{\mathcal{D}}

\newcommand{\calL}{\mathcal{L}}
\newcommand{\calT}{\mathcal{T}}

\newcommand{\calI}{\mathcal{I}}

\begin{document}

\title{Collaborative Multi-Teacher Knowledge Distillation for Learning Low Bit-width Deep Neural Networks}

\author{
Cuong Pham\textsuperscript{\rm 1}
\and
Tuan Hoang\textsuperscript{\rm 2}
\and
Thanh-Toan Do\textsuperscript{\rm 1}
\and
\textsuperscript{\rm 1}Department of Data Science and AI, Monash University, Australia \\
\textsuperscript{\rm 2}Bytedance \\
{\tt\small van.pham1@monash.edu},  {\tt\small tuan.hoang@bytedance.com},  {\tt\small toan.do@monash.edu}
}

\maketitle
\thispagestyle{empty}


\begin{abstract}
   Knowledge distillation which learns a lightweight student model by distilling knowledge from a cumbersome teacher model is an attractive approach for learning compact deep neural networks (DNNs). Recent works further improve student network performance by leveraging multiple teacher networks. However, most of the existing knowledge distillation-based multi-teacher methods use separately pretrained teachers. This limits the collaborative learning between teachers and the mutual learning between teachers and student. Network quantization is another attractive approach for learning compact DNNs. However, most existing network quantization methods are developed and evaluated without considering multi-teacher support to enhance the performance of quantized student model. In this paper, we propose a novel framework
   that leverages both multi-teacher knowledge distillation and network quantization for learning low bit-width DNNs. The proposed method encourages both collaborative learning between quantized teachers and mutual learning between quantized teachers and quantized student. During learning process, at corresponding layers, knowledge from teachers will 
   form an importance-aware shared knowledge which will be used as input for teachers at subsequent layers and also be used to guide student. 
  Our experimental results on CIFAR-100 and ImageNet datasets show that the compact quantized student models trained with our method achieve competitive results compared to other state-of-the-art methods, and in some cases, 
  surpass the full precision models. 
   
\end{abstract}

\section{Introduction}
\label{sec:intro}
Deep Convolutional Neural Networks (CNNs) have achieved tremendous successes in a variety of computer vision tasks, including image classification, object detection, segmentation, and image retrieval. However, CNNs generally require excessive memory and expensive computational resources, limiting their usage in many applications. Therefore, a great number of researches have been devoted to {making} CNNs lightweight and to {improving} inference efficiency for practical applications.

An effective approach is to use low bit-width weights and/or low bit-width activations. This approach not only can reduce the memory footprint but also achieves a significant gain in speed, as the most computationally expensive convolutions can be done by bitwise operations \cite{xnor_net}. Although existing quantization-based methods \cite{HWGQ,lossaware,xnor_net,tbn,LQNets,dorefa,TTQ} achieved improvements, there are still noticeable accuracy gaps between the quantized CNNs and their full precision counterparts, especially in the challenging cases of 1 or 2 bit-width weights and activations. 

Model compression using knowledge distillation is another attractive approach to reduce the computational cost of DNNs \cite{overhaul_fea_distill,KD_Hinton,QKD,apprentice,FitNets}. In knowledge distillation, a smaller student network is trained to mimic the behaviour of a cumbersome teacher network. 
To further enhance the performance of {the} student network, some works~\cite{adaptive_multi_teacher,DBLP:conf/ecai/ParkK20,DBLP:conf/kdd/YouX0T17} propose to distill knowledge from multiple teachers. However, in those works, teacher models are separately pretrained, which would limit the collaborative learning between teachers. It also limits the mutual learning between student network and teacher networks.  

To improve the compact low bit-width student network, in this paper, we propose a novel framework -- Collaborative Multi-Teacher Knowledge Distillation (CMT-KD), which encourages the {collaborative learning} between teachers and the mutual learning between teachers and student. 

In the collaborative learning process, knowledge at corresponding layers from teachers will be combined to form an importance-aware shared knowledge which will subsequently be used as input for the next layers of teachers.
The collaborative learning between teachers is expected to form a valuable shared knowledge to be distilled to the corresponding layers in the student network. To our best knowledge, this paper is the first one that proposes this kind of collaborative learning for knowledge distillation.

It is worth noting that our novel framework design allows end-to-end training, in which not only the teachers and student networks but also
the contributions (i.e., importance factors) of teachers to the shared knowledge are also learnable during the learning process. It is also worth noting that the proposed framework is flexible -- different quantization functions and different knowledge distillation methods can be used in our framework.

To evaluate the effectiveness of the proposed framework, we conduct experiments on CIFAR-100 and ImageNet datasets with AlexNet and ResNet18 architectures. The results show that the compact student models trained with our framework achieve competitive results compared to previous works.

\section{Related work}
\label{sec:related_work}
Our work is closely related to two main research topics in the literature: network quantization and knowledge distillation (KD). 

\textbf{Network quantization.} 
Earlier works in network quantization have applied the basic form of weight quantization to directly constrain the weight values into the binary/ternary space without or with a scaling factor, i.e., $\{-1, 1\}$  \cite{BinaryConnect}, $\{-\alpha, \alpha\}$  \cite{xnor_net}, or $\{-\alpha, 0, \alpha\}$ \cite{TWN}. 
Since quantization of activations can substantially reduce complexity further \cite{xnor_net,BinaryNet,LQNets,dorefa}, this research topic attracts more and more attention \cite{xnor_net,LQNets,dorefa}. 
In~\cite{xnor_net,BinaryNet}, the authors propose to binarize both weights and activations to $\{-1,1\}$. However, there are considerable accuracy drops compared to full precision networks. 
{To address this problem, the generalized low bit-width quantization \cite{log_data_represent,dorefa} is 
studied.
}
{In half-wave Gaussian quantization (HWGQ) \cite{HWGQ}, the authors propose a practical and simple uniform quantization method that exploits the statistics of network activations and batch normalization.}
In LQ-Nets \cite{LQNets}, the authors propose to  train a quantized CNN and its associated non-uniform quantizers jointly.
The approach in Learned Step Size Quantization (LSQ) \cite{LSQ} learns uniform quantizers using trainable interval values. In quantization-interval-learning (QIL) \cite{QIL}, the authors introduce a trainable quantizer that additionally performs both pruning and clipping.

\textbf{Knowledge Distillation (KD)} is a common method in training smaller networks by distilling knowledge from a large teacher model \cite{KD_Hinton}. The rationale behind this is to use extra supervision in the forms of classification probabilities \cite{KD_Hinton}, intermediate feature representations~\cite{mimic,ABdistill,overhaul_fea_distill,FitNets}, attention maps \cite{attention_transfer}.
Knowledge distillation approaches transfer the knowledge of a teacher network to a student network in two different settings: offline and online. In the offline learning, KD uses a fixed pre-trained teacher network to transfer the knowledge to the student network. Deep mutual learning \cite{DML} mitigates this limitation by conducting online distillation in one-phase training between two peer student models.

\textbf{Multi-Teacher Knowledge Distillation.}
The approach in \cite{MultiLang_NMT} applies multi-teacher learning into multi-task learning where each teacher corresponds to a task. Similarly, the approach in \cite{UHC} trains a classifier in each source and unifies their classifications on an integrated label space.
The approach in \cite{DBLP:conf/kdd/YouX0T17} considers knowledge from multiple teachers equally by averaging the soft targets from different pretrained teacher networks. In
\cite{adaptive_multi_teacher}, the authors propose to learn a weighted combination of pretrained teacher representations.  
{Different from \cite{DBLP:conf/kdd/YouX0T17,adaptive_multi_teacher}, in this work, we propose a novel online distillation method that captures importance-aware knowledge from different teachers 
before distilling the captured knowledge to the student network.} {
In \cite{DBLP:conf/ecai/ParkK20}, the last feature map of student network is fed through different non-linear layers; each non-linear layer is for each teacher. The student network and the non-linear transformations are trained such that the output of those non-linear transformations mimic the last feature maps of the corresponding teacher networks.}
The previous works  \cite{adaptive_multi_teacher,DBLP:conf/ecai/ParkK20, DBLP:conf/kdd/YouX0T17} mainly learn full precision student models from a set of full precision pretrained teacher models, while we focus on learning quantized models. Specifically, we aim to learn a quantized student model with guidance from a set of quantized teacher models with different precisions. 
In addition, different from previous works \cite{DBLP:conf/kdd/YouX0T17,adaptive_multi_teacher,DBLP:conf/ecai/ParkK20} in which teachers are fixed when training student, our method simultaneously trains student and teachers using the collaborative and mutual learning.

\textbf{Quantization + Knowledge distillation}. Some works have tried to adopt knowledge distillation methods to assist the training process of low precision networks \cite{SPEQ,QKD,apprentice,distill_quant,guidedQuantize}. 
In Apprentice (AP) \cite{apprentice}, the teacher and student networks are initialized with the corresponding pre-trained full precision networks. After lowering the precision of the student network, the student is then fine-tuned using distillation.
Due to AP’s initialization of the student, AP might get stuck in a local minimum in the case of very low bit-width quantized student networks. 
Because of the inherent differences between the feature distributions of the full-precision teacher and low-precision student network, using a fixed teacher as in \cite{apprentice} can limit the knowledge transfer. QKD \cite{QKD} and Guided-Quantization \cite{guidedQuantize} mitigate the issue of AP by jointly training the teacher and student models, which makes a teacher adaptable to the student model.
In our work, to further mitigate the problem of different {feature distributions} between teacher and student models, instead of using a full-precision teacher model, we propose to use a set of quantized teacher models. Using quantized models would help the teachers obtain more suitable knowledge for a quantized student model to mimic.

\section{Proposed method}
\label{sec:proposed}

\subsection{The proposed framework}
\begin{figure*}[!h]
\centering
  \includegraphics[scale=1.8, width=0.9\linewidth]{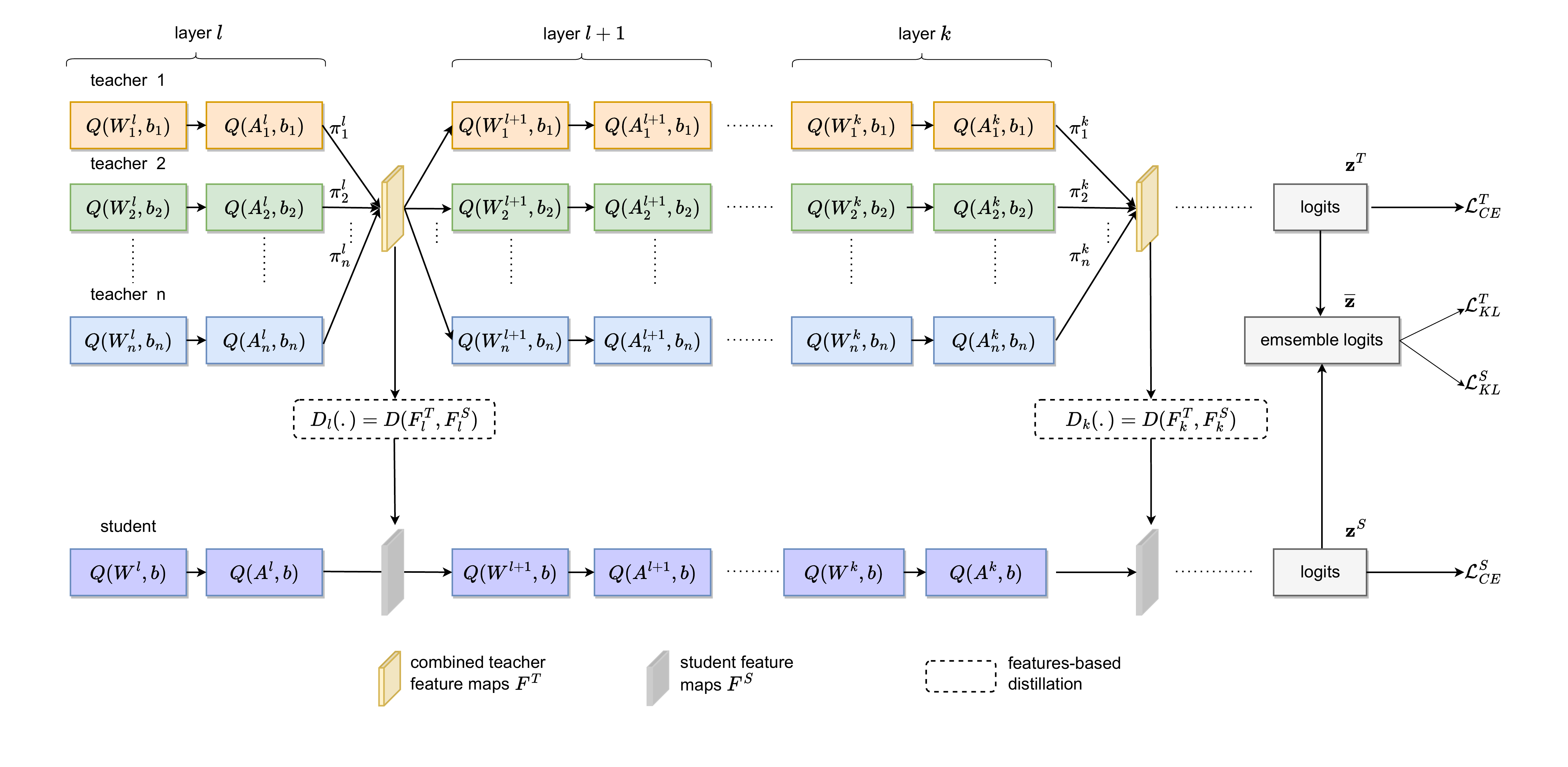}
    \caption{The framework of our proposed collaborative multi-teacher knowledge distillation for low bit-width DNNs. 
    The  collaborative  learning among a set of quantized teachers forms useful shared knowledge at certain layers through importance parameters $(\pi)$. 
    $\calL^T_{KL}$ and $\calL^S_{KL}$ are for mutual learning between teachers and student 
    via ensemble logits $\overline{\bz}$, where $\overline{\bz}$ is calculated from teachers logits $\bz^T$ and student logits $\bz^S$. 
    CE means Cross Entropy, and KL means Kullback-Leibler divergence. $\calD(.)$ denotes the loss for intermediate features-based distillation that could be attention loss or hint loss (FitNet). }
\label{fig:1}       
\vspace{-0.5em}
\end{figure*}

We propose a novel framework -- Collaborative Multi-Teacher Knowledge Distillation (CMT-KD) as illustrated in Figure~\ref{fig:1}, which encourages collaborative learning between teachers and mutual learning between teachers and student. 

First, we propose a novel collaborative learning of multiple teachers. During the training process, the  collaborative  learning among a set of quantized teachers forms useful importance-aware shared knowledge at certain layers, 
which is distilled  to  the  corresponding  layers in the student network. 
Second, the learning process between the student and the teachers is performed in a mutual-learning manner \cite{DML} via ensemble logits $\overline{\bz}$ from teachers logits $\bz^T$ and student logits $\bz^S$. 


Furthermore, our framework design allows end-to-end training, in which not only the teachers and student networks but also the contributions (i.e., importance factors) of teachers to the shared knowledge are also learnable during the learning process. It is also worth noting that the proposed framework is flexible -- different quantization functions \cite{HWGQ,HWGQ_Rethinking,LSQ,LQW_CAQ,QIL,LQNets} and different knowledge distillation methods \cite{mimic,ABdistill,overhaul_fea_distill,FitNets,attention_transfer} can be used in our framework.

\subsection{Collaborative learning of multiple teachers}
Teacher model selection is important in knowledge distillation. In~\cite{QKD}, the authors show that if there is a large gap in the capacity between teacher and student, the knowledge from teacher may not be well transferred to student. To control the power of teacher, in our work, we consider teacher and student models that have the same architecture. However, teachers are quantized with higher different bit-widths. 
The knowledge at corresponding layers from teachers will be fused to form a  shared knowledge which will subsequently be used as input for the next layers of teachers. This forms collaborative learning between teachers. It is expected that different teachers have different capacities, and therefore, they should have different contributions to the shared knowledge. To this end, for each teacher, an importance factor that controls how much knowledge the teacher will contribute to the shared knowledge will also be learned. It is worth noting that the learning of importance factors will encourage the collaborative learning between teachers to produce a suitable shared knowledge that the student can effectively mimic.


Formally, given a quantization function $Q(x, b)$,  
the $i^{th}$ teacher is quantized using $Q(W_i, b_i)$ and $Q(A_i, b_i)$, in which $W_i$ and $A_i$ represent for weights and activations, respectively; $b_i$ is the bit-width.  
The shared knowledge $F_k$ of corresponding layers of $n$ teachers at $k^{th}$ layer index is formulated as follows

\begin{equation}
\begin{gathered}
F_k = \sum_{i=1}^{n}\pi^{k}_i * Q(A_i^k, b_i) \\
     \text { s.t. } \sum_i \pi^{k}_i =1, \pi^{k}_i \in[0,1] \text {,}
\end{gathered}
\label{eq:Fk}
\end{equation}
where 
${\pi^{k}_i}$ represents the importance of teacher $i^{th}$. To handle the constraint over $\pi$ in (\ref{eq:Fk}), in the implementation, a softmax function is applied to $\pi^{k}_i$ values before they are used to compute $F_k$.
The importance parameters $\pi$ and the model weights of teachers and student are optimized simultaneously via end-to-end training.
\subsection{Other components}

\subsubsection{Quantization function}
Quantization function maps a value $x \in R$ to a quantized value $\overline{x} \in \{q_1, q_2, ..., q_n\}$ using a quantization function Q with a precision bit-width $b$. The quantized value is defined as
\begin{equation}
    \overline{x}=Q(x, b).
\end{equation}


Different quantization methods have been proposed \cite{HWGQ,LSQ,QIL,xnor_net,LQNets,dorefa}. In this paper, we consider the half-wave Gaussian quantization (HWGQ) \cite{HWGQ} as a quantizer to be used in our framework, which is an effective and simple uniform quantization approach. 
To quantize weights and activations, they first pre-compute the optimal value $q_i$ by using uniform quantization {for unit Gaussian distribution}. 
{Depending on variance $\sigma$ of  weights and activations}, the quantized value of $x$ is expressed as
\begin{equation}
    \overline{x} = \sigma*q_i.
\end{equation}


\subsubsection{Intermediate features-based distillation}

The shared knowledge from teachers will be used to guide the learning of student. 
Let $F^T_k$ and $F^S_k$ be the shared feature map of teachers and the feature map of the student at $k^{th}$ layers of models, respectively. Let $\calI$ be the selected layer indices for intermediate features-based distillation.
The intermediate features-based knowledge distillation loss is defined as follows


\begin{equation}
    \calL_{feat} = \sum_{k \in \calI} \calD \left(F_{k}^{T}, F_{k}^{S}\right),
    \label{eq:fea_loss}
\end{equation}
where $\calD$ is the distance loss measuring the similarity between features $F^T_k$ and $F^S_k$.
Different forms of $\calD$ can be applied in our framework. In this work, we consider two widely used distance losses, i.e., the attention loss~\cite{attention_transfer} and the FitNet loss\cite{FitNets}. 



The attention loss~\cite{attention_transfer} is defined as follows
\begin{equation}
    \calD_{A T}=\sum_{k \in \calI}\left\|\frac{Q_{k}^{T}}{\left\|Q_{k}^{T}\right\|_{2}}-\frac{Q_{k}^{S}}{\left\|Q_{k}^{S}\right\|_{2}}\right\|_{p},
\end{equation}
where $Q_k^S$ and $Q_k^S$ are the attention maps of features $F_k^S$ and $F_k^S$, respectively. $\|.\|_p$ is the $l_p$ norm function that could be a $l_1$ or $l_2$ normalization. The attention map for a feature map $F \in R^{c \times h \times w}$ that has $c$ channels is defined as $Q=\sum_{j=1}^{c}|A_j|^p$, where  $A_j = F(j, :,:)$ is the $j^{th}$ channel of $F$; $|.|$ is the element-wise absolute function. In our implementation we use $p=2$. 
The FitNet loss \cite{FitNets} is defined as follows 
\begin{equation}
    \calD_{HT}=\sum_{k \in \calI}\left\|F_{k}^{T} - r(F_{k}^{S})\right\|_{p},
\end{equation}
where $r(.)$ is a convolutional layer that adapts the student feature map $F_k^{S}$ before comparing it to the shared knowledge teacher feature map $F_k^{T}$. In our method, we follow existing works \cite{overhaul_fea_distill,FitNets} in which $r(.)$ a convolutional layer with a kernel size of $1\times1$.
\subsubsection{Mutual learning between teachers and student}
In addition to the intermediate features-based distillation, we also leverage the mutual learning \cite{DML} defined on the logits from networks for learning. The mutual learning allows student to give feedback about its learning to teachers. This learning mechanism encourages both teachers and student to simultaneously adjust their parameters to achieve an overall learning objective, e.g., minimizing a total loss function.

The mutual learning \cite{DML} applies $KL$ losses on the softmax outputs of networks.
However, due to the diversity of the output logits from different networks, this method may hurt the performance of models. 
To overcome this issue, we adopt KDCL-MinLogit \cite{KDCL}, which  is a simple and effective method to ensemble logits from teachers and student. In particular, this method selects the minimum logit value of each category. 

Let $\bz^T$ and $\bz^S$ be the logit outputs of the combined
teacher model $T$ and a student model $S$, $z^{T,c}$ and $z^{S,c}$ be the elements of $\bz^T$ and $\bz^S$ corresponding to the target class $c$, $\mathbf{1}$ be the vector with all $1s$ elements, we denote $\bz^{T, c} = \bz^T - z^{T,c}\mathbf{1}$ and $\bz^{S,c} = \bz^S - z^{S,c}\mathbf{1}$. 
The element $\overline{z_i}$ of the ensemble logits $\overline{\bz}$ is computed as follows
\begin{equation}
    \overline{z_i} = min\{z_i^{T,c},  z_i^{S,c}\}, i=1, 2, ..., m
\end{equation}
where  $z_i^{T,c}$, $z_i^{S,c}$ are the $i^{th}$ elements of $\bz^{T, c}$ and $\bz^{S, c}$, and $m$ is the number of classes.


The mutual learning is defined as follows
%

\begin{equation}
    \calL_{KL}^{S}= \calT^{2} \times KL(\overline{\bp}|| \bp^{S}),
\end{equation}
\begin{equation}
    \calL_{KL}^{T}= \calT^{2} \times KL(\overline{\bp}|| \bp^{T}),
\end{equation}
where $KL$ means the Kullback-Leibler  divergence and $\calT$ is the temperature parameter. $\overline{\bp}$, $\bp^{S}$, and $\bp^{T}$ are the soft logits of $\overline{\bz}$, $\bz^{S}$, and $\bz^{T}$, respectively. The soft logit $\bp$ is defined as $\bp=softmax(\frac{\bz}{\calT})$.


Finally, the overall loss of our proposed collaborative and mutual learning in classification task is defined as

\begin{equation}
\small
\label{eq:final_loss}
    \begin{aligned}
    \calL = \alpha \times (\calL_{CE}^{S}+ \calL_{CE}^{T}) 
    + \beta \times (\calL_{KL}^S + \calL_{KL}^T) + \gamma \times \calL_{feat},
\end{aligned}
\end{equation}
where $\alpha$, $\beta$, and $\gamma$ are the hyper-parameters of total loss for optimization and $\calL_{CE}$ is the standard cross-entropy loss calculated using the corresponding soft logits and the ground-truth labels. During the training process, 
both the model weights of teachers, student and the importance factors of teachers, i.e., $\pi_i^k ~\forall i,k$, will be updated by minimizing $\calL$, using gradient descent. 

\section{Experiments}
\label{sec:experiments}
\subsection{Experimental setup}
\paragraph{\textbf{Datasets.}}
We conduct experiments on {CIFAR-100} \cite{cifar} and {ImageNet} (ILSVRC-2012) \cite{imagenet} datasets. CIFAR-100 dataset consists of $100$ classes with the total of $60,000$ images, where $50,000$ and $10,000$ images are used for training and testing set, respectively. ImageNet is a large scale dataset with $1,000$ classes in total. This dataset contains $1.2$ million images for training and $50,000$ images for validation, which is used as test set in our experiments.

\paragraph{\textbf{Implementation details.}} 
We evaluate our proposed method on two common deep neural networks AlexNet \cite{alexnet} and ResNet18 \cite{Resnet}. Regarding AlexNet, batch normalization layers are added after each convolutional layer and each fully connected layer, which are similar to works done by \cite{HWGQ, dorefa}. In all experiments, similar to previous works \cite{LQNets,guidedQuantize}, in the training, we use the basic augmentation including horizontal flips, resizing 
and randomly cropping that crops images to $227\times 227$ and $224\times 224$ pixels for ResNet18 and AlexNet, respectively. 
We use Stochastic Gradient Descents with a momentum of $0.9$ and a mini-batch size of $256$. The learning rate $lr$ for network models is set to $0.1$ and $0.01$ for ResNet and AlexNet, respectively. 
The learning rate for the importance factors ($\pi$) in teacher models is set to $lr/10$.


When training ResNet18 model on CIFAR-100, we train the model with 120 epochs. The learning rate is decreased by a factor of 10 after 50 and 100 epochs. When training ResNet18 model on ImageNet, we train the model with 100 epochs. The learning rate is decreased by a factor of 10 after $30$, $60$, and $90$ epochs. When training AlexNet model, for both CIFAR-100 and ImageNet, we train the model with $100$ epochs and we adopt $cosine$ learning rate decay.

We set the weight decay to 25e-6 for the 1 or 2-bit precision and set it to 1e-4 for higher precisions. Regarding hyper-parameters of the total loss (\ref{eq:final_loss}), {we empirically set $\alpha=1$, $\beta=0.5$}. 
In our experiments, the shared knowledge is formed at certain layers of teachers and is distilled to the correspondence layers of student. 
Specifically, the shared knowledge is formed at the last convolutional layers of each convolution block, i.e., layers $2, 5$, and $7$ of AlexNet teachers and from layers $5, 9, 13$, and $17$ of  ResNet18 teachers. 
Meanwhile, we set $\gamma$ to $100$ or $1$ when attention loss or FitNet loss is used in $\calL_{feat}$. We do not quantize the first and last layers. 








\subsection{Ablation studies}
\label{subsec:ablation}

We conduct several ablation studies on CIFAR-100 with ResNet18 and AlexNet to demonstrate the effectiveness of our proposed method.
For ablation studies, we use HWGQ quantizer \cite{HWGQ} for the proposed CMT-KD. In addition, for the intermediate features-based distillation ($\calL_{feat}$) in the final loss (eq. (\ref{eq:final_loss})), the attention loss is used. We consider the following settings. 

\begin{table}[!t]
\def\arraystretch{1.0}
  \centering
  \begin{tabular}{c|c|c|c}
    \hline
    Models & Bit-width & Top 1 & Top 5 \\
    \hline
    \multirow{5}{*}{Single model} 
     & FP   & 72.4 & 91.3 \\
     & 8 bits& 70.9 & 90.9 \\
     & 6 bits& 70.8 & 90.8 \\
     & 4 bits& 70.7 & 90.8 \\
     & 2 bits& 69.4 & 90.5 \\
     \hline
    KD (from FP teacher) & \multirow{5}{*}{2 bits} & 
     {71.3} & {91.6} \\
    Average teacher & & 71.0 & 91.6 \\
    
    CMT-KD (w/o Att) &  & 71.8 & 91.8 \\
    CMT-KD (w/o ML) &  & 70.9 & 91.3 \\
    CMT-KD &  & 72.1 & 91.9 \\
     \hline
    Combined teacher &4, 6, 8 bits  & 72.3 & 91.7 \\
    
  \hline
  \end{tabular}
   \caption{Ablation studies on the CIFAR-100 dataset with AlexNet. The descriptions of settings are presented in Section \ref{subsec:ablation}.}
    \label{tab:ablationAlexNet} 
\end{table}

\begin{table}[!t]
\def\arraystretch{1}
  \centering
  \begin{tabular}{c|c|c|c}
   \hline
    Models & Bit-width & Top 1 & Top 5 \\
    \hline
    \multirow{5}{*}{Single model} 
    &FP& 75.3 & 93.1 \\
     &8 bits& 75.2 & 92.9 \\
     &6 bits&  74.9 & 92.7 \\
     &4 bits& 74.9 & 92.5 \\
     &2 bits& 72.9 & 91.9 \\
     \hline
    KD (from FP teacher) & \multirow{5}{*}{2 bits} & {75.1} & {92.8} \\
    Average teacher & & 76.0 & 93.8 \\
    
    CMT-KD (w/o Att) &  & 76.5 & 93.9 \\
    CMT-KD (w/o ML) &  & 75.0 & 93.2 \\
    CMT-KD &  & {78.3} & {94.4} \\
    \hline
    Combined teacher &4, 6, 8 bits  & 79.5 & 94.9 \\
     
    \hline
  \end{tabular}
   \caption{Ablation studies on the CIFAR-100 dataset with ResNet18. The descriptions of settings are presented in Section \ref{subsec:ablation}.}
    \label{tab:ablationResNet18} 
    \vspace{-0.5em}
\end{table}

    
\paragraph{\textbf{Single model.}} We evaluate single models with different precisions (i.e., full precision, 8-bit, 6-bit, 4-bit, and 2-bit precisions) without any distillation methods. The results for AlexNet and ResNet18 architectures are presented in Table \ref{tab:ablationAlexNet} and Table \ref{tab:ablationResNet18}, respectively. With the AlexNet architecture, the results show that the  4-bit, 6-bit, 8-bit models achieve comparable results. There are considerable large gaps between these models and the full precision model. There are also large gaps between those models and the 2-bit model. With the ResNet18 architecture, the 8-bit model achieves comparable results to the full precision model. There is a small gap between the 6-bit, 4-bit models and the 8-bit model. Similar to the observation on the AlexNet, there is a large gap between the 2-bit model and the full precision model. 
\paragraph{\textbf{Knowledge distillation from the full precision teacher.}} In this setting, we train a 2-bit student model with the knowledge distillation from the full precision teacher, i.e., the KD (from FP teacher) setting in Table \ref{tab:ablationAlexNet} and Table \ref{tab:ablationResNet18}. For this setting, we follow \cite{KD_Hinton}, i.e., when training the student, in addition to the cross-entropy loss, the softmax outputs from the teacher will be distilled to the student. When AlexNet is used, this setting achieves better performance than quantized single models. When ResNet18 is used, this setting achieves comparable results to the 8-bit single quantized model. 

\paragraph{\textbf{Knowledge distillation from an ensemble of multiple quantized teachers.}}
In this setting, we separately train three teachers with different precisions, i.e., 4-bit, 6-bit, and 8-bit teachers. The averaged softmax outputs of teachers are distilled to the 2-bit student. This setting is noted as ``Average teacher" in Table \ref{tab:ablationAlexNet} and Table~\ref{tab:ablationResNet18}. It is worth noting that this setting is also used in the previous work \cite{DBLP:conf/kdd/YouX0T17}.
 When AlexNet is used, at the top-1 accuracy, this setting produces the 2-bit student that achieves comparable performance with the quantized single teachers. 
 However, its performance ($71.0\%$) is still lower than the full precision model ($72.4\%$). When ResNet18 is used, this setting improves the performance over the full precision model, i.e., the gain is $0.7\%$ for both top-1 and top-5 accuracy.  

\paragraph{\textbf{Effectiveness of collaborative and mutual learning.}} We consider different settings of the proposed framework.  
For the results in Table \ref{tab:ablationAlexNet} and Table \ref{tab:ablationResNet18}, when training CMT-KD models, we use 3 teachers, i.e., 4-bit, 6-bit, and 8-bit teachers. In those tables, CMT-KD means that the models are trained with the total loss (\ref{eq:final_loss}). CMT-KD (w/o Att) means that the models are trained with the loss (\ref{eq:final_loss}) but the intermediate features-based component $\calL_{feat}$ is excluded. CMT-KD (w/o ML) means that the  models are trained with the loss (\ref{eq:final_loss}) but the mutual learning component $(\calL_{KL}^S + \calL_{KL}^T)$ is excluded. The results show that the mutual learning loss is more effective than the intermediate features-based loss. However, both components are necessary to achieve the best results, i.e., CMT-KD. 

When AlexNet is used, the full precision (FP) model slightly outperforms the proposed CMT-KD at top-1 accuracy. 
When ResNet18 is used, CMT-KD outperforms the FP model at both top-1 and top-5 accuracy. A significant gain is achieved, i.e., $3\%$, at top-1 accuracy. It is worth noting that when ResNet18 is used, the CMT-KD significantly outperforms the 2-bit single model, the 2-bit model when using the average teacher, and the 2-bit model when distilling from the FP model. Those results confirm the effectiveness of the proposed method. 

\paragraph{\textbf{Combined teacher.}} We also evaluate the performance of the combined teacher in the collaborative learning in our proposed method, i.e., the predictions which are made by the classifier corresponding to the $\calL^T_{CE}$ loss in Figure \ref{fig:1}. Overall, this setting produces the best results, except for top-1 accuracy with AlexNet architecture. It achieves better  performance than the ``Average teacher'' setting. 
With ResNet18, this setting significantly outperforms the full precision model. Those results confirm the effectiveness of the proposed collaborative learning between teachers. 
\begin{table}[]
\vspace{-1em}
\vspace{0.5em}
\def\arraystretch{1.1}
\centering
    \begin{tabular}{c|ccc}
    \hline Setting & & AlexNet & ResNet18 \\
    \hline \multirow{2}{*}{ (a) } & Top-1 & {72.1} & {78.3} \\
        & Top-5 & 91.9 & 94.4 \\
    \hline \multirow{2}{*}{ (b) } & Top-1 & 71.1 & 78.1 \\
                                  & Top-5 & 91.2 & 94.3 \\
    \hline
    \end{tabular}
\vspace{-0.5em}
\caption{Impact of the number of teachers on the CMT-KD 2-bit students. The results are on the CIFAR-100 dataset. (a) Using 4-bit, 6-bit, and 8-bit teachers. (b) Using 4-bit and 8-bit  teachers.}
\label{tab:numteachers}
\end{table}

\paragraph{\textbf{Impact of the number of teachers.}} The results in Table \ref{tab:numteachers} show the impact of the number of teachers on the performance of the 2-bit CMT-KD student models. The results show that using 3 teachers (4-bit, 6-bit, and 8-bit) slightly improves the performance when using 2 teachers (4-bit and 8-bit).

\subsection{Comparison with the state of the art}
In this section, we compare our proposed method CMT-KD against state-of-the-art network quantization methods, including LQ-Net \cite{LQNets}, LQW+CAQ \cite{LQW_CAQ}, HWGQ \cite{HWGQ}, and DoReFa-Net \cite{dorefa}. We also make a comparison between our approach and methods that apply both distillation and quantization consisting of PQ+TS+Guided \cite{guidedQuantize}, QKD \cite{QKD}, SPEQ \cite{SPEQ}. 
For CMT-KD, we use three teachers (4-bit, 6-bit, and 8-bit teachers) to guide the learning of compact quantized 2-bit weights ($K_w = 2$) and 2-bit activations ($K_a = 2$) students. Meanwhile, we use 2-bit, 4-bit, and 8-bit teachers to guide the learning of compact quantized 1-bit weights ($K_w = 1$) and 2-bit activations ($K_a = 2$) students.
We do not consider  1-bit activations because the previous \cite{HWGQ,NIPS2016_d8330f85,xnor_net,dorefa} show that 1-bit quantization for activations is not sufficient for good performance. 
\setlength{\tabcolsep}{4pt}
\begin{table}[!t]   
\def\arraystretch{1.05}
\centering
\vspace{0.5em}
\begin{tabular}{l|c|c|c|c}
\hline 
\multirow{2}{*}{ Method }  & \multicolumn{2}{|c|}{ AlexNet } & \multicolumn{2}{c}{ ResNet18 } \\
\cline{2-5}   & Top-1 & Top-5 & Top-1 & Top-5  \\
\hline 
\hline
\multicolumn{1}{l|}{Full precision} & 72.4 & 91.3 & 75.3 & 93.1  \\
\hline 
\multicolumn{5}{c}{ $K_w=1$, $K_a=2$ } \\
\hline
LQ-Nets \cite{LQNets}  & 68.7 & 90.5 & 70.4 & 91.2 \\
LQW + CAQ \cite{LQW_CAQ}  & 69.3 & 91.2 & 72.1 & 91.6 \\
HWGQ \cite{HWGQ}  & 68.6 & 90.8 & 71.0 & 90.8 \\
{CMT-KD-FitNet}   & 69.9 & \textbf{91.3} & \textbf{76.1} & \textbf{93.7} \\
  {CMT-KD-Att}   & \textbf{70.4} & {91.1} & {75.6} & {93.5} \\
\hline 
\multicolumn{5}{c}{ $K_w=2$, $K_a=2$ } \\
\hline
PQ+TS+Guided \cite{guidedQuantize}  & 64.6 & 87.8 & - & - \\
LQ-Net \cite{LQNets}  & 69.2 & 91.2 & 70.8 & 91.3 \\
LQW + CAQ \cite{LQW_CAQ}  & 69.9 & {91.3} & 72.1 & 91.6 \\
HWGQ  \cite{HWGQ}  & 69.4 & 90.5 & 72.9 & 91.9 \\
{CMT-KD-FitNet}   & {70.0} & {90.7} & \textbf{78.7} & \textbf{94.6} \\
{CMT-KD-Att} &\textbf{72.1} & \textbf{91.9} & 78.3 & 94.4 \\
\hline
\end{tabular}
\caption{The comparative results on the CIFAR-100 dataset. We report the results of CMT-KD when FitNet loss or attention loss (Att) is used for the intermediate features-based distillation. CMT-KD uses HWGQ quantizer to quantize teachers and student. The results of HWGQ are reported by using the official released code.}
\label{tab:sota_cifar10} 
\end{table}
\paragraph{\textbf{Comparative results on CIFAR-100.}} Table \ref{tab:sota_cifar10} presents the top-1 and top-5 classification accuracy on CIFAR-100 dataset of different network quantization methods for AlexNet and ResNet18. {The results of competitors are cited from \cite{LQW_CAQ,guidedQuantize}}. 
We report the results of our CMT-KD when FitNet loss or attention loss is used for $\calL_{feat}$ in Eq. (\ref{eq:final_loss}). Those models are denoted as CMT-KD-FitNet or CMT-KD-Att. The quantizer HWGQ \cite{HWGQ} is used to quantize teachers and student networks when training CMT-KD models. 
Overall, the best CMT-KD models outperform most of the competitor quantization methods. When AlexNet is used, 
the CMT-KD (for both FitNet and Att) models outperform the compared quantization methods at top-1 accuracy. However, the proposed models achieve lower performance than the FP model at top-1 accuracy. This may be due to the limit in the capacity of the AlexNet model, which consists of only 5 convolutional layers.  

When ResNet18 is used, our CMT-KD models outperform the full precision model. Especially when using 2-bit weights and 2-bit activations, the improvements of the CMT-KD-FitNet over the FP model are $3.4\%$ and $1.5\%$ for top-1 and top-5 accuracy, respectively. 
{It is also worth noting that the proposed models significantly improve over the HWGQ method~\cite{HWGQ} which uses HWGQ quantizer to quantize FP models, i.e., with $K_w=2, K_a = 2$, CMT-KD-FitNet outperforms HWGQ \cite{HWGQ} $5.8\%$ at top-1 accuracy.} 

\begin{table}[!t]
\def\arraystretch{1.05}
\centering
\vspace{0.5em}
\begin{tabular}{l|c|c|c|c}
\hline 
\multirow{2}{*}{ Method }  & \multicolumn{2}{|c|}{ AlexNet } & \multicolumn{2}{c}{ ResNet18 } \\
\cline{2-5}   & Top-1 & Top-5 & Top-1 & Top-5  \\
\hline 
\hline
\multicolumn{1}{l|}{ Full precision} & 61.8 & 83.5 & 70.3 & 89.5  \\
\hline 
\multicolumn{5}{c}{ $K_w=1$, $K_a=2$ } \\
\hline
DoReFa-Net \cite{dorefa} & 49.8 & - & 53.4 & - \\
LQ-Nets \cite{LQNets}   & 55.7 & 78.8 & \textbf{62.6} & \textbf{84.3} \\
HWGQ \cite{HWGQ}   & 52.7 & 76.3 & 59.6 & 82.2 \\
{CMT-KD (HWGQ)}   &\textbf{56.2}  &\textbf{79.1}  & 60.6 & 83.5 \\
\hline 
\multicolumn{5}{c}{ $K_w=2$, $K_a=2$ } \\
\hline 
DoReFa-Net \cite{dorefa}  & 48.3 & 71.6 & 57.6 & 80.8  \\
QKD \cite{QKD}   & - & - & 67.4 & 87.5  \\
PQ + TQ Guided \cite{guidedQuantize}   & 52.5 & 77.3 & - & -  \\
LQ-Net \cite{LQNets}   & 57.4 & 80.1 & 64.9 & 85.9  \\
SPEQ \cite{SPEQ}   & 59.3 & - & 67.4 & -  \\
HWGQ \cite{HWGQ,HWGQ_Rethinking}   & 58.6 & 80.9 & 65.1 & 86.2  \\
CMT-KD (HWGQ)  &59.2  &81.3  & {65.6} & {86.5} \\
LSQ (with distill) \cite{LSQ}    & -  & -    & \textbf{67.9} & \textbf{88.1}  \\
LSQ* (w/o distill)  & -  & -    & {66.7} & {87.1}  \\
CMT-KD (LSQ)   &\textbf{59.3}  &\textbf{81.5}  & {67.8} & {87.8} \\
\hline
\end{tabular}
\caption{The comparative results on the ImageNet dataset. CMT-KD uses the attention loss for the intermediate features-based distillation. CMT-KD (HWGQ) and CMT-KD (LSQ) denote models when HWGQ \cite{HWGQ} and LSQ \cite{LSQ} quantizers are used in our framework, respectively. We report experimental results for LSQ (please refer to footnote 1) without distillation in LSQ* row.} 
\label{tab:sota_imagenet} 
\end{table}
\vspace{-0.5em}
\paragraph{\textbf{Comparative results on ImageNet.}}

\begin{figure*}[!t]
     \centering
     \begin{subfigure}[b]{0.37\textwidth}
         \centering
         \includegraphics[width=\textwidth]{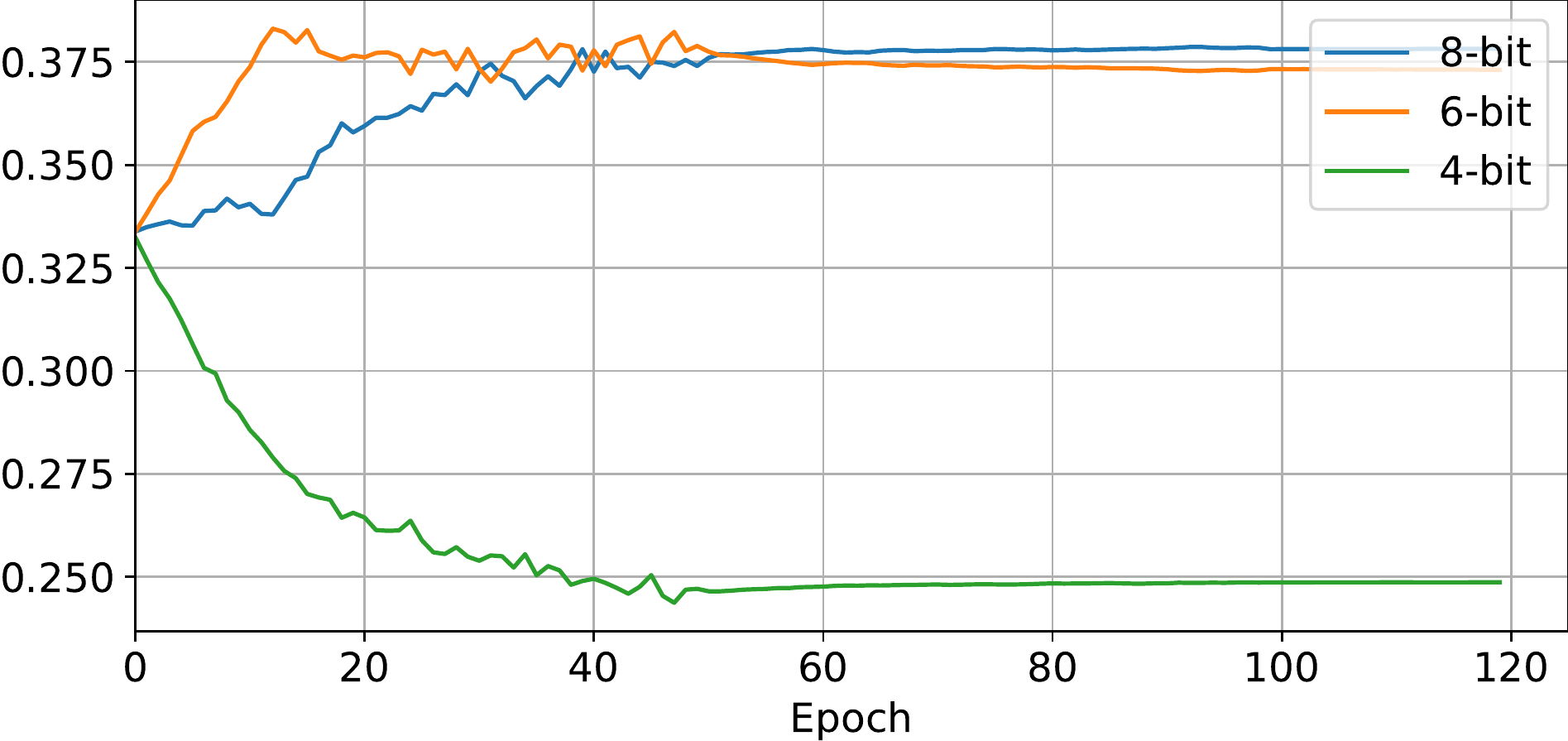}
         \caption{Layer 5}
         \label{fig:layer 1}
     \end{subfigure}
     \hspace{0.8cm}
    \begin{subfigure}[b]{0.37\textwidth}
         \centering
         \includegraphics[width=\textwidth]{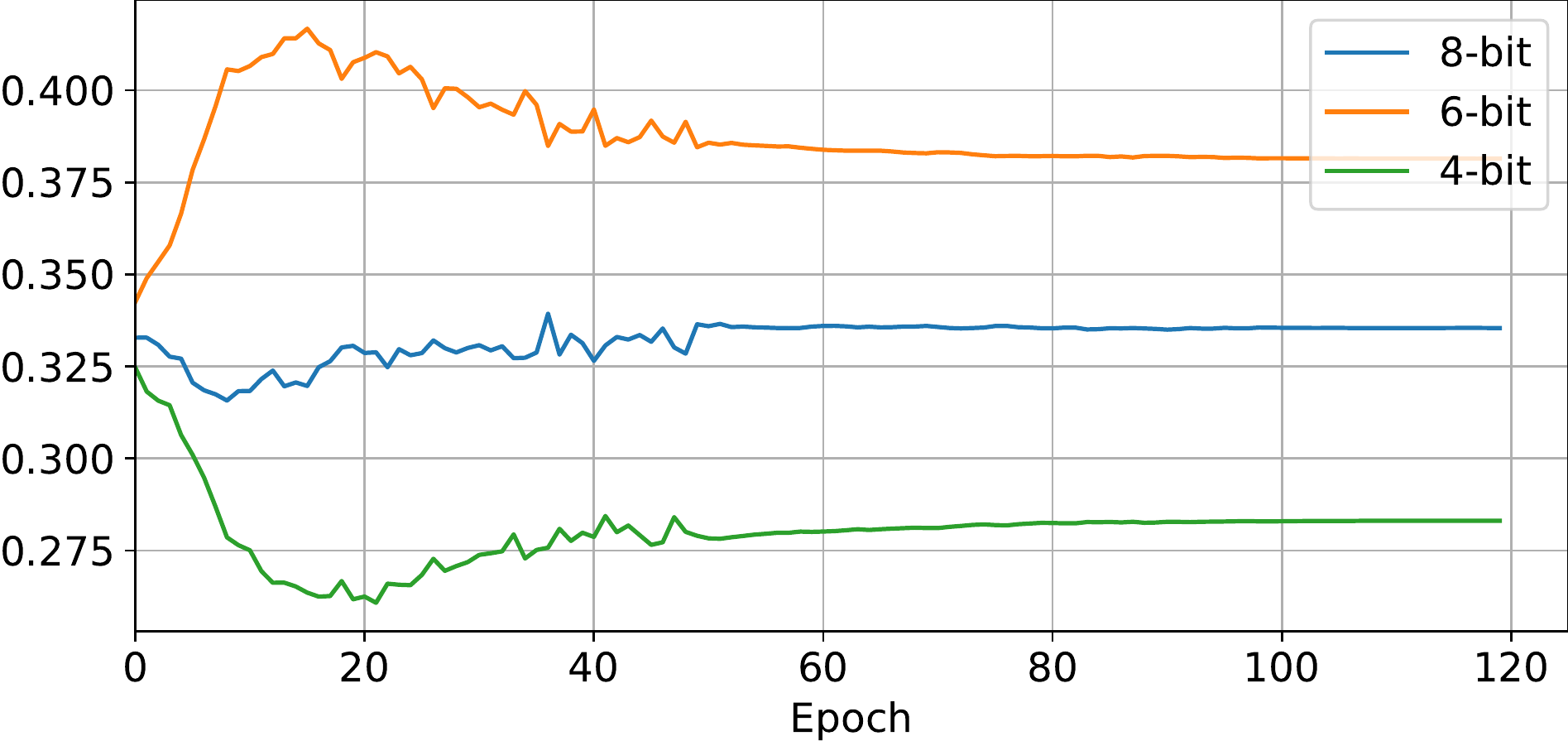}
         \caption{Layer 9}
         \label{fig:layer 2}
     \end{subfigure}
     
     \begin{subfigure}[b]{0.37\textwidth}
         \centering
         \includegraphics[width=\textwidth]{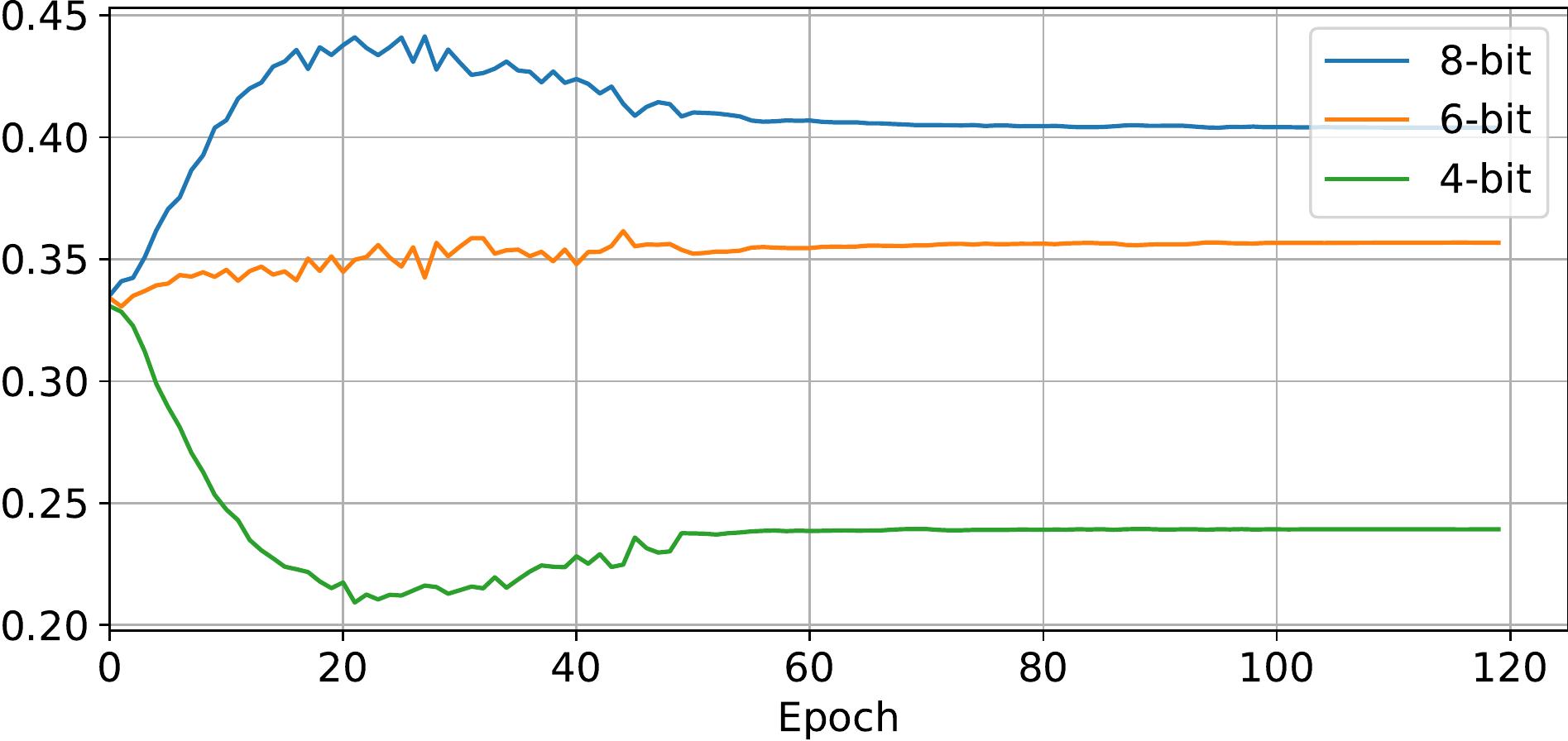}
         \caption{Layer 13}
         \label{fig:layer 3}
     \end{subfigure}
     \hspace{0.8cm}
     \begin{subfigure}[b]{0.37\textwidth}
         \centering
         \includegraphics[width=\textwidth]{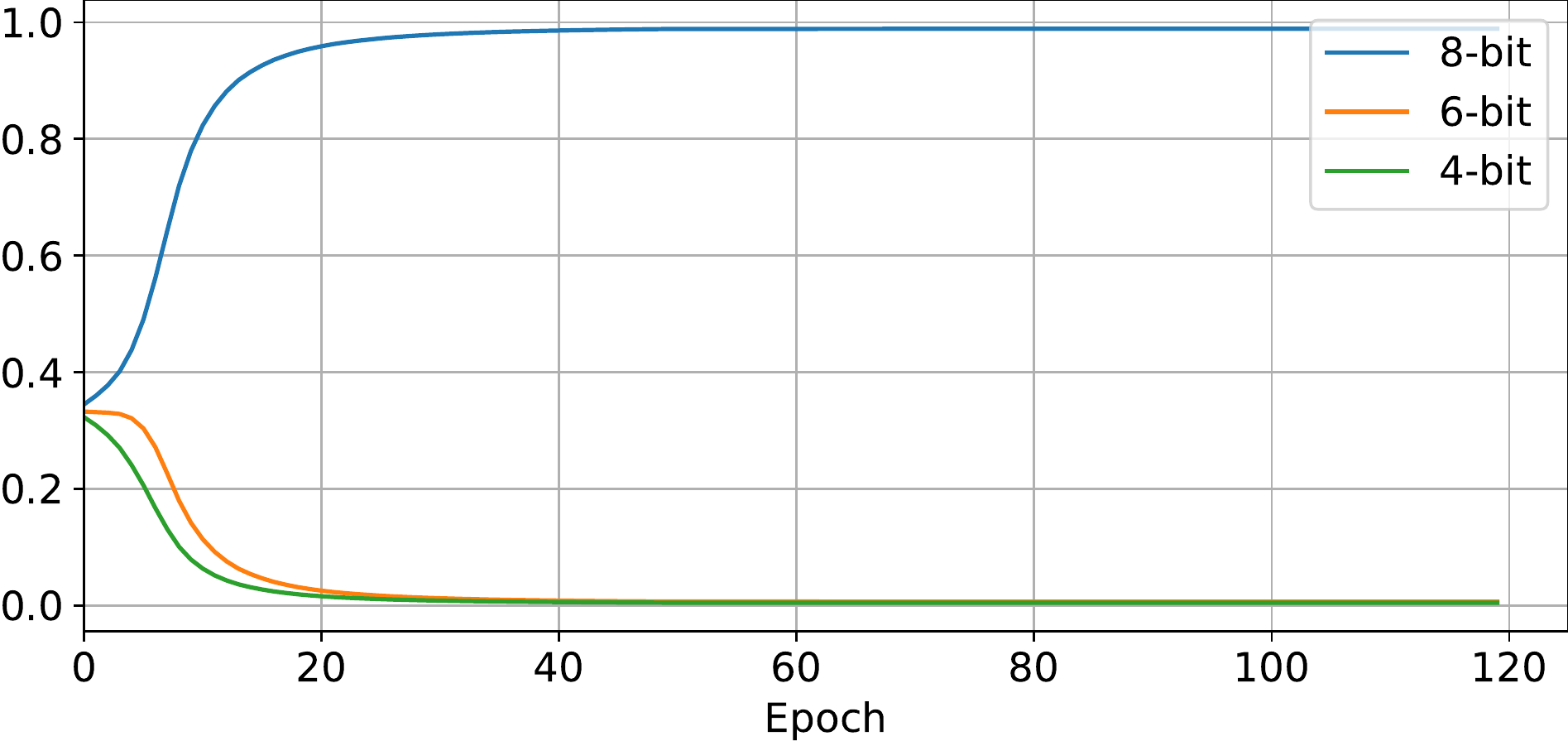}
         \caption{Layer 17}
         \label{fig:layer 4}
     \end{subfigure}
        \caption{The importance factors of the three ResNet18 teachers (4-bit, 6-bit, 8-bit) on the CIFAR-100 dataset during the training process.}
        \label{fig:importancefactor}
        \vspace{-1em}
\end{figure*}

Table \ref{tab:sota_imagenet} presents the top-1 and top-5 classification accuracy on ImageNet dataset of different network quantization methods for AlexNet and ResNet18. The results of competitors are cited from the corresponding papers. 
At $K_w = 1$ and $K_a =2$, when using AlexNet, the proposed CMT-KD  significantly outperforms HWGQ \cite{HWGQ}. The gain is $4.5\%$ and $2.8\%$ for top-1 and top-5 accuracy, respectively. For ResNet18, we also achieved an improvement of $1\%$ compared to HWGQ at top-1 accuracy. 
With $K_w = 2$ and $K_a =2$, the CMT-KD (HWGQ)  outperforms the HWGQ \cite{HWGQ} method by $0.6\%$, and $0.5\%$ on top-1 accuracy for AlexNet and ResNet18, respectively. 

As the proposed framework is flexible to quantizers, we also report in Table \ref{tab:sota_imagenet} the results when LSQ~\cite{LSQ} quantizer is used to quantize teacher and student networks in our framework, i.e., CMT-KD (LSQ). LSQ is a quantization method in which the step size is learned during training. 
%
%
 When ResNet18 is used, given the same LSQ quantizer implementation\footnote{\label{note1}The official source code of LSQ is not available. We adopt the LSQ quantizer from an un-official implementation from \url{https://github.com/hustzxd/LSQuantization} for our experiments.}, our method CMT-KD (LSQ) can boost the top-1 accuracy of LSQ* by $1.1\%$. 
However, we note that the results reported in LSQ \cite{LSQ} slightly outperforms  CMT-KD (LSQ). It is worth noting that in order to achieve the reported results ($67.9\%$ top-1 and $88.1\%$ top-5), LSQ \cite{LSQ} also uses knowledge distillation to distill knowledge from the full precision model to the quantized model\footnote{
We are unsuccessful in reproducing results reported in LSQ \cite{LSQ}. For example, without distillation for LSQ, we only achieve a result of $66.7\%$ for top-1 accuracy when using ResNet18 with $K_w=2, K_a=2$, while in \cite{LSQ}, the authors reported $67.6\%$ at the same setting.}. 
Our best method CMT-KD (LSQ) compares favorably to the recent method SPEQ~\cite{SPEQ} for both AlexNet and ResNet18 models.
\paragraph{\textbf{Visualization of the importance factors ($\pi$) of teachers.}}
Figure \ref{fig:importancefactor} visualizes the importance factors of three teachers (4-bit, 6-bit, 8-bit) when using ResNet18 architecture for the teachers and student. The experiment is conducted on CIFAR-100 dataset. The visualized importance factors are at the last convolutional layers in each block of ResNet18 during training, i.e., layers 5, 9, 13, 17. They are also the layers in which the shared knowledge is formed. 
Figure \ref{fig:layer 2} shows that the highest precision teacher (8-bit) does not always give the highest contributions to the shared knowledge at all layers. For example, at layer 9, the importance factor of the 6-bit teacher is higher than the importance factor of the 8-bit teacher. At the last convolutional layer (i.e., layer 17), the 8-bit teacher dominates other teachers and gives most of the contributions to the shared knowledge.  In addition, the importance factors are mainly updated at the early training stage of the framework.  



\section{Conclusion}
\label{sec:conclusion}
In this paper, we propose a novel approach for learning low bit-width DNNs models by distilling knowledge from multiple quantized teachers. We introduce the idea of collaborative learning that allows teachers to form importance-aware shared knowledge, which will be used to guide the student. The proposed framework also leverages the idea of mutual learning that allows both teachers and student to adjust their parameters to achieve an overall object function. The proposed framework allows end-to-end training in which not only network parameters but also the importance factors indicating the contributions of teachers to the shared knowledge are updated simultaneously. The experimental results on CIFAR-100 and ImageNet datasets with AlexNet and ResNet18 architectures demonstrate that the low bit-width models trained with the proposed approach achieve competitive results compared to the  state-of-the-art methods. 

{\small
\bibliographystyle{ieee_fullname}
\bibliography{cmt_kd}
}
\end{document}